# Outlier Guided Optimization of Abdominal Segmentation

Yuchen Xu[*a], Olivia Tang[*a], Yucheng Tang[**a], Ho Hin Lee[a], Yunqiang Chen[b], Dashan Gao[b],
Shizhong Han[b], Riqiang Gao[a], Michael R. Savona[c], Richard G. Abramson[d],
Yuankai Huo[a], Bennett A. Landman[a,d]

[a]Department of Electrical Engineering and Computer Science, Vanderbilt University, Nashville, TN, USA 37212;
[b]12 Sigma Technology, San Diego, CA, USA 92130;
[c]Hematology and Oncology, Vanderbilt University Medical Center, Nashville, TN, USA 37235
[d]Radiology, Vanderbilt University Medical Center, Nashville, TN, USA 37235
(*Equal contribution **Corresponding author: yucheng.tang@vanderbilt.edu)

## ABSTRACT

Abdominal multi-organ segmentation of computed tomography (CT) images has been the subject of extensive research interest. It presents a substantial challenge in medical image processing, as the shape and distribution of abdominal organs can vary greatly among the population and within an individual over time. While continuous integration of novel datasets into the training set provides potential for better segmentation performance, collection of data at scale is not only costly, but also impractical in some contexts. Moreover, it remains unclear what marginal value additional data have to offer. Herein, we propose a single-pass active learning method through human quality assurance (QA). We built on a pre-trained 3D U-Net model for abdominal multi-organ segmentation and augmented the dataset either with outlier data (e.g., exemplars for which the baseline algorithm failed) or inliers (e.g., exemplars for which the baseline algorithm worked). The new models were trained using the augmented datasets with 5-fold cross-validation (for outlier data) and withheld outlier samples (for inlier data). Manual labeling of outliers increased Dice scores with outliers by 0.130, compared to an increase of 0.067 with inliers ($p<0.001$, two-tailed paired t-test). By adding 5 to 37 inliers or outliers to training, we find that the marginal value of adding outliers is higher than that of adding inliers. In summary, improvement on single-organ performance was obtained without diminishing multi-organ performance or significantly increasing training time. Hence, identification and correction of baseline failure cases present an effective and efficient method of selecting training data to improve algorithm performance.

**Keywords:** computed tomography, deep convolutional neural networks, multi-organ segmentation, abdomen segmentation, active learning

## 1. INTRODUCTION

With the rapid increase in the availability of computational power to medical research, there has been more focus on acquiring large bodies of data in an effort to increase the performance of various algorithms, including medical image segmentation. However, labeled data are precious resources and can be extremely costly to obtain [1]. The focus of this work is on evaluating potential strategies for expending resources to create labeled data and understanding the relative tradeoffs between potential labeling options.

Active learning has been shown in other domains to outperform "passive learning" (based on random sampling) by requiring a smaller number of annotated samples to achieve a similar level of performance [2]. It also has the advantage of detecting infrequent classes and improving a classifier with a lower cost of labeling [3]. Proposed methods for active learning by other researchers include training multiple generative adversarial networks to generate a distribution of outlier data [4] and distance-based outlier detection [3]. The concept of outlier mitigation is the primary motivation of this work as we posit that global failures ("outliers") in a small fraction of testing subjects may lead to systems that cannot be reliably deployed in practice.

We propose a single-pass active learning method through human quality assurance (QA) to optimize abdomen multi-organ segmentation. Specifically, we are interested in studying the marginal benefits of adding new subjects to a baseline segmentation algorithm that has achieved satisfactory performance on the majority of cases but suffers from rare/infrequent outliers. We hypothesize that manually detecting outliers (Fig. 1B), addressing them, and using them to augment the training data provides an efficient and effective way of improving the overall outcome of the algorithm. The pancreas was selected as a sample organ because its shape and size are highly variable among the population, which makes it crucial to the success of multi-organ segmentation tasks. We added manually labeled outlier data incrementally and contrasted the results to adding machine-predicted inlier data (where the baseline algorithm did well on; Fig. 1A). We also visually examined examples of multi-organ predictions generated by the baseline and improved algorithms.

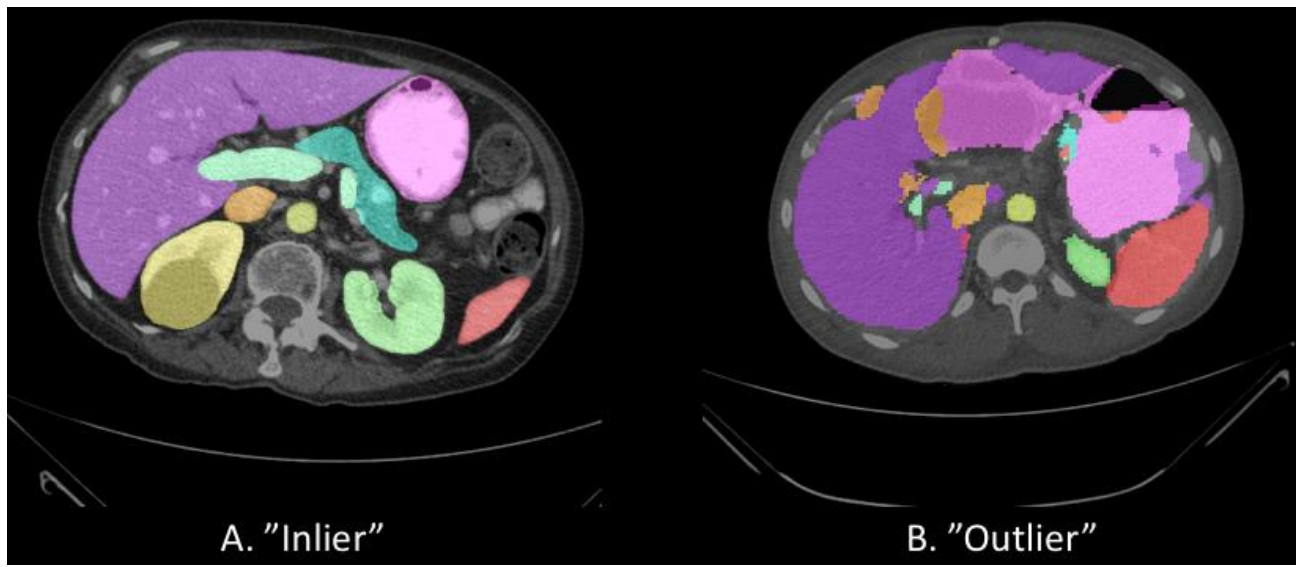

Figure 1. Examples of segmentations generated by the baseline algorithm. A. An inlier, where the algorithm correctly predicted larger organs, like the liver (on the left side of image in purple) and spleen (on the right side of image in red-pink) and suggested mostly accurate areas of smaller organs. B. An outlier (global failure), where the liver (purple) and spleen (red-pink) were largely correct but major inconsistencies are visible with other organs.

## 2. METHODS

### 2.1 Preparatory Work and Data

#### 2.1.1 Baseline Data and Algorithm

A total of 220 abdomen CT scans of dimensions (512, 512, ~120) with pixel sizes of (0.8mm, 0.8mm, 2.5mm), enhanced to the portal venous phase, was used in training a 3D U-Net model [5]. 100 scans were manually segmented all-organ images from representative cases of deidentified patient data [6], with a 80%-20% training-validation split (Group 1A and 1B in Fig. 2, accordingly). The other 120 subjects were deidentified splenomegaly patients with manually labeled spleens [7], all of which were used for the training set (Group 2). A separate clinical dataset of 6,317 deidentified clinical scans were used for assessing the baseline algorithm. All data were retrieved in deidentified form under IRB approval.

#### 2.1.2 Preprocessing

The abdomen region of CT scans was obtained by running a body-part regression algorithm through whole-body scans [8]. All scans used in the experiment were soft-tissue windowed at minima of -175 and maxima of 275 Hounsfield units. The images were then down-sampled to pixel sizes of (2mm, 2mm, 6mm) and dimensions (160, 160, 64) and cropped or padded to trainable sizes.

#### 2.1.3 Manual QA

The pre-trained baseline model was evaluated on the 6,317 deidentified clinical scans, and manual QA was performed on the predictions generated by the algorithm in a simple Python GUI software. Predictions were separated into three discrete groups: above state-of-the-art (where the algorithm achieved satisfactory performance; "success"/"inliers"), usable (where segmentations were mostly correct but lacking in details), and global failures ("failure"/"outliers").

#### 2.1.4 Manual Labeling of Outlier Data

There was a total of 817 failed scans, 47 of which were manually labeled on the pancreas (Group 3).

Manual tracing of the pancreas was performed in the Medical Image Processing, Analysis, and Visualization (MIPAV) application version 8.0.2 by the National Institute of Health [9]. Tracing was mostly completed in the axial plane, checked for consistency of shape in the coronal and sagittal planes. The spine was first located, followed by the aorta, the inferior vena cava, the superior mesenteric artery, and the superior mesenteric vein (SMV) to its anterior side. Meanwhile, the portal vein (PV) from the liver and the splenic vein (SV) from the spleen were identified and followed across axial slices where SMV, PV, and SV meet. The pancreas was traced under a soft tissue window as the deformable structure above and surrounding the splenic vein with smooth and plentiful curvature, which could be confirmed in the coronal slices. An average of one hour per scan was observed for labeling the pancreas.

### 2.1.5 Inlier Data

A group of 37 inliers (Group 4) (out of 598) was randomly selected to parallel the number of outliers. Note that Group 4 is smaller in size than Group 3, because of cross-validation used in Experiment 1, detailed in Section 2.2.4. The predicted areas of the pancreas generated by the baseline algorithm were used as labels of the pancreas.

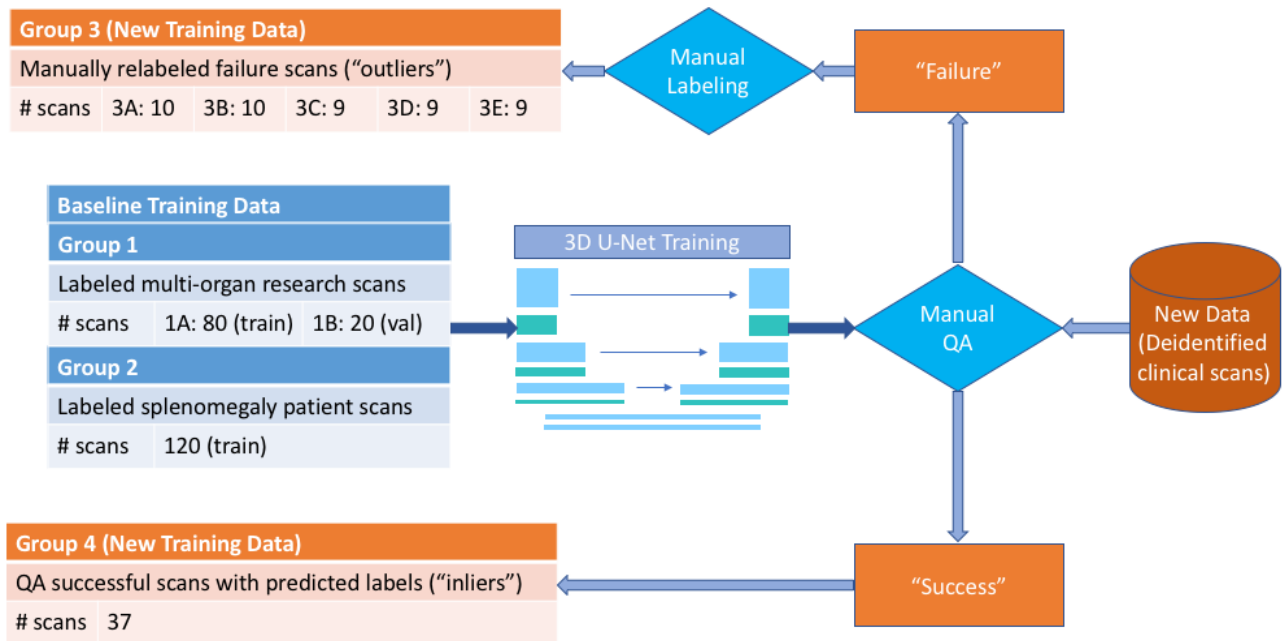

Figure 2. Overview of data employed in the experiment. To train the baseline 3D U-Net model [5], 200 scans from Groups 1A and 2 were used in the training set and 20 from Group 1B were used in the validation set. The resulting model was then applied to a separate clinical dataset, and the outputs were evaluated using manual QA. Specifically, major failures ("outliers") and successes ("inliers") were identified and went to Groups 3 and 4, accordingly. Manual labeling was used to correct the failures. Scans in Group 3 were divided into 5 sub-groups 3A through 3E for use in cross-validation in Experiment 1, as described in Section 2.2.4 and Table 1.

## 2.2 Experimental Design

### 2.2.1 Metric of Evaluation

The Dice similarity coefficient, or Dice score, was used to evaluate the segmentation results against ground truth (manual labels) since a higher Dice score typically indicates better segmentation performance. During training, to maintain that the loss decreases with time, the additive inverse of the Dice score was used to calculate the loss.

### 2.2.2 Transfer Learning

The model pre-trained on the baseline training data was used for initialization in each experiment. The same learning rate 0.0001, as in baseline training, was adopted.

### 2.2.3 Validation Set

A group of 20 out of the 100 manually segmented all-organ images from the baseline training data were used in the validation set, in addition to 10% of the newly added scans in each experiment (rounded up; the remaining scans were used in the training set). When the average Dice score on the pancreas across the validation set did not increase by more than 0.001 in 10 epochs, the epoch that produced the highest Dice score by far was selected as the model for use in testing.

Table 1. Experimental design with description of data. The training and validation sets used for the baseline algorithm remained in place (200 scans in training and 20 in validation). A fraction of scans from Group 3 or Group 4 were added to the sets of Experiments 1 and 2, accordingly, using a 90%-10% training-validation split. The 47 scans from Group 3 were used for testing. To ensure that the testing scans were fully withheld in Experiment 1, a 5-fold cross-validation was used, detailed in Section 2.2.4 (X=5, 10, 15, 20, 25, 30, 35, 37).

| Datasets | Experiment 1: **Adding manually labeled outliers** | Experiment 2: **Adding QA successful scans (inliers)** |
|---|---|---|
| Training Set | 80 (Group 1A) + 120 (Group 2) + 90%X (Group 3; rounded down) | 80 (Group 1A) + 120 (Group 2) + 90%X (Group 4; rounded down) |
| Validation Set | 20 (Group 1B) + 10%X (Group 3; rounded up) | 20 (Group 1B) + 10%X (Group 4; rounded up) |
| Testing Set | 47 (**5-fold cross-validation**) (Group 3) | 47 (Group 3) |

### 2.2.4 Experiment 1: Augmenting the Dataset with Outlier Data

In experiment 1, the models were trained on outliers (Group 3) added in increments of 5, from 5 to 35 (and a maximum at 37), in addition to the 220 scans from the baseline data (Table 1). Note that all 47 outliers should not be added at once because at least one sub-group in Group 3 should be withheld for testing at any time. Part of the data was used for the validation sets as described in 2.2.3.

Because the 47 manually labeled outliers were also used for testing, 5-fold cross-validation [10] was used in this experiment for each outlier scan to be fully withheld exactly once during training. Specifically, in the first fold, Group 3A was withheld for testing, while scans were randomly selected from Groups 3B through 3E, which were then divided into training and validation sets. In the second fold, Group 3B was withheld for testing, and scans were randomly selected from Groups 3A, 3C, 3D and 3E.

Therefore, for each number of outliers (X) added, there were 5 models (1 from each fold) that were tested on 5 sub-groups of the outliers (3A through 3E), accordingly. A weighted average (weighted on the number of scans in each sub-group) was taken to obtain the testing results for each X.

### 2.2.5 Experiment 2: Augmenting the Dataset with Inlier Data

In experiment 2, the models were trained on inliers (Group 4) added in increments of 5, from 5 to 35 (and a maximum at 37), in addition to the 220 scans from the baseline data. Part of the data was used for the validation sets as described in 2.2.3. Each number of inliers added resulted in one model, which was then evaluated on the 47 outliers (Group 3).

### 2.2.6 Assessment of Results

The testing results were visualized using a boxplot showing outliers, emphasizing the comparison between the outcomes of adding inlier and outlier scans. We hypothesize that adding outlier scans results in better performance on the testing set than adding inlier scans and performed two-tailed paired t-tests between the same images in the testing set to assess the hypothesis. Performance of optimized models was also evaluated qualitatively through visual inspection of predicted multi-organ segmentations (as opposed to single-organ segmentations focused on the pancreas) to characterize the global impact of manually labeling an important organ.

## 3. RESULTS

As shown in Fig. 3, the baseline algorithm has an average Dice score of 0.4196 (std=0.211) on the pancreas. Models trained with added inlier subjects produced Dice scores between 0.4536 (5) and 0.4881 (35), while those trained with outlier subjects produced between 0.5047 (5) and 0.5500 (37). All new models performed significantly better than the baseline model (two-tailed paired t-test, p<0.05 for adding 5 inliers, p<0.001 for adding 15 inliers, p<0.0001 for others) All outlier-augmented models did significantly better than their inlier counterparts (see Fig. 3 for p-values). Slight increases are observed within the same experiment with more subjects added to the training set. All groups except one showed various numbers of outlier data on the lower end, defined by subjects with Dice scores more than one interquartile range lower than the lower quartile.

Examples of images with Dice scores at 25%, 50% and 75% percentile in the baseline algorithm were selected and shown in Fig. 4, along with multi-organ segmentation predictions from the newly trained models.

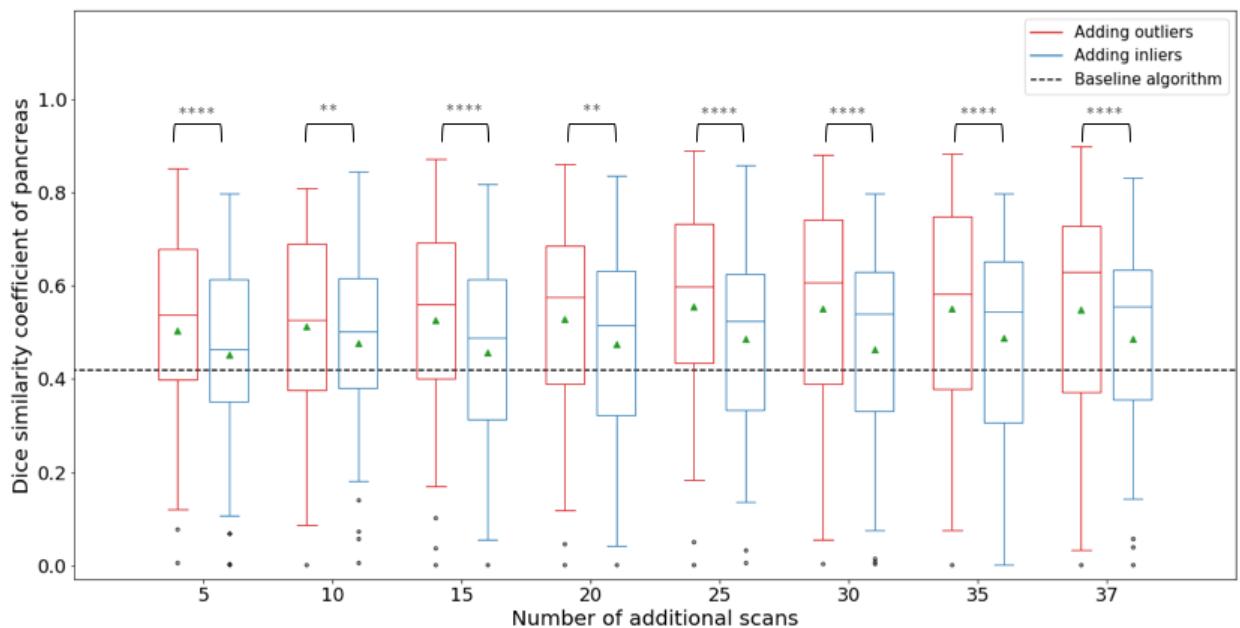

Figure 3. Dice scores on the pancreas of testing scans. Red boxplots on the left show the results of adding outlier subjects and blue boxplots on the right inlier subjects, compared with the baseline algorithm (denoted by the dark dashed line). Green triangles show the means of each group, lines in the middle of the box the medians, and black dots outliers of scores. The whiskers extend up to one interquartile range above and below the upper and lower quartile. There was also a weak correlation between number of subjects added and the Dice scores on testing data: for inlier-augmented models, Pearson correlation coefficient=0.08, $r^2$=0.47, p=0.06; for outlier-augmented models, Pearson correlation coefficient=0.16, $r^2$=0.77, p<0.005. (**p<0.01; ****p<0.0005; two-tailed paired t-test on scores of the same testing scans between outlier-augmented and inlier-augmented models)

## 4. DISCUSSION

From Figure 3, all newly trained models achieved higher Dice scores than the baseline model by a substantial amount. This confirms that the transfer learning method we used, in which a model pre-trained by a subset of the new training data was used for initialization of the experiment, was effective. Transfer learning was also highly efficient in this case, where it took an average of 6.5 epochs for validation results to achieve optima, taking as low as 1-3 epochs in some cases and not exceeding 20. Segmentation results also showed a reasonable amount of consistency across the three models as shown in Fig. 4.

Segmentation performance from adding outlier subjects is significantly better than from adding inliers, as all eight groups shown in Fig. 3 have high confidence in the differences. The differences seem to be slightly more robust with more subjects added. It is worth noting, however, that the differences are modest in size, which could be interpreted as the size effect in play: the baseline training set contains 100 images with the pancreas labeled, and the 37 subjects added should be considered a significant increase in size. Still, adding more outliers showed a meaningful (albeit modest) influence at scale while adding inliers did not.

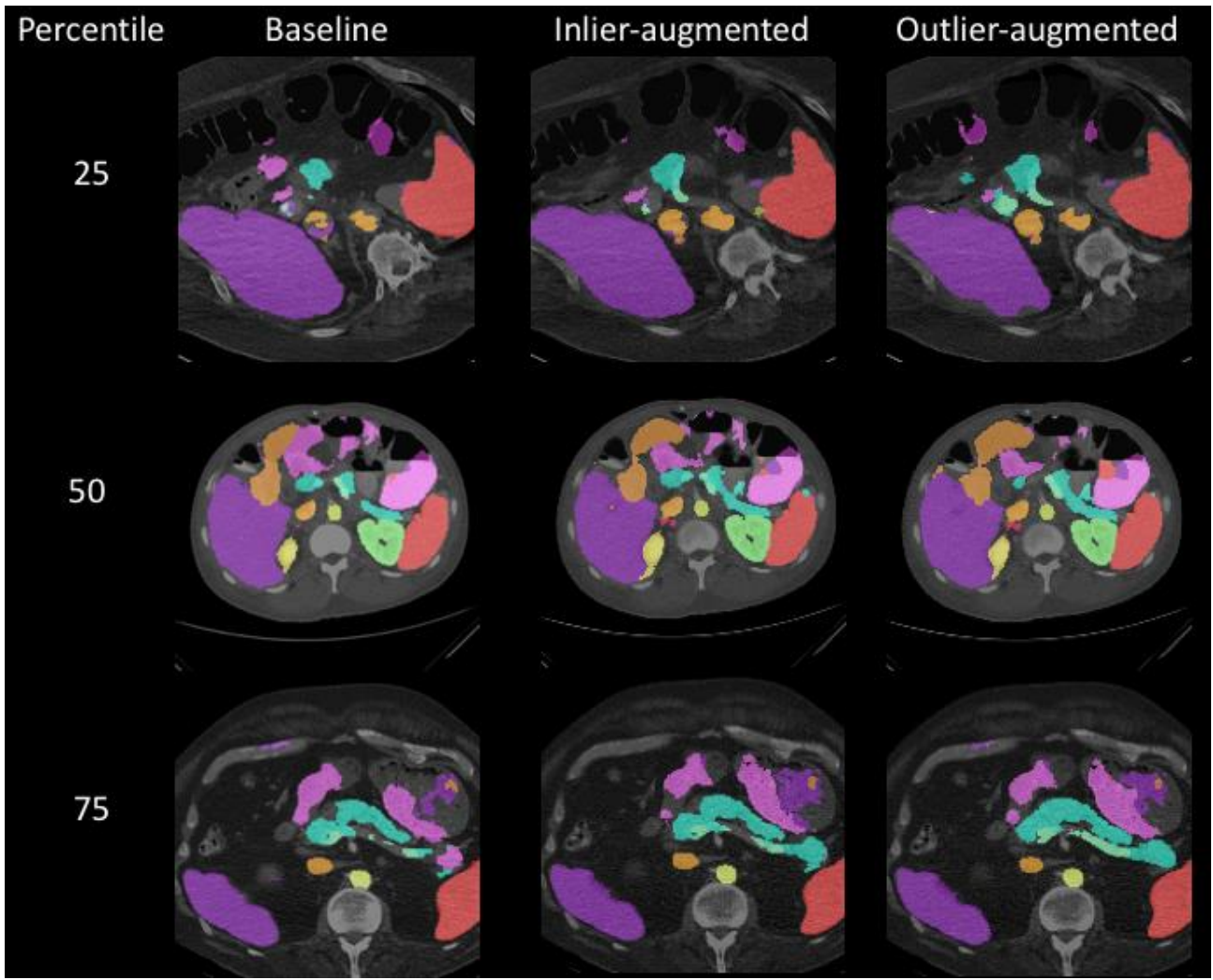

Figure 4. Comparison of performance between the three models on examples. The three images had Dice scores on the pancreas at 25%, 50% and 75% percentile in the baseline model predictions, accordingly, and the same slices of the same images predicted by newly trained models with added inlier and outlier data were shown in parallel. All images are in axial planes, and different colors represent different organs. Our target organ, the pancreas, is in blue.

Visually, in Fig. 4, the new models do not appear to be much better than the baseline model, and there does not seem to be a major difference between the inlier-augmented and outlier-augmented outcomes. Such similarity is reasonable considering that neither a Dice score of 0.40 nor a 0.55 on the pancreas are ideal compared to the ~0.7 state-of-the-art accuracy [11, 12], although it is still worth noting that new models, especially the outlier-augmented one, show visible improvements on segmentation results with other organs. This is important because outlier-guided optimization on a single organ may not only preserve multi-organ results but has potential to improve the overall performance.

Curiously, all but one experimental group have outlier scores on the lower end in Fig. 3. Those images are the same ones that also had highly inaccurate results with the baseline algorithm, and they do not appear to improve in training. Considering recent work on active learning [2], it could be inferred that abdominal segmentation could also be divided into numerous "classes" of images, and learning more classes improves the overall performance of the model. Detecting and adding labeled outlier data is more efficient in learning infrequent classes, but if outliers in the one set are not clustered, they are not helpful (i.e., multiple instances of outliers from the same "class" need to be added for them to have a practical impact).

In summary, this work showed that active learning by human detection of outliers, manual labeling, and augmenting the training set with them show promising potential for improving abdominal segmentation performance both on single-organ and multi-organ tasks effectively and efficiently. Future research could address several insufficiencies in this work. In terms of the dataset, the training-validation split was uneven with the multi-organ and pancreas training data, which may have resulted in bias towards the former set; furthermore, no completely hold-out dataset was used for testing due to the limited amount of data available. In terms of experimental design, a mixture of inliers and outliers, or a completely random sample of subjects could also be used to augment the dataset to further evaluate the argument made in this paper.

## 5. ACKNOWLEDGEMENTS

This research is supported by Vanderbilt-12Sigma Research Grant, NSF CAREER 1452485 and NIH 1R01EB017230 (Landman). This study was in part using the resources of the Advanced Computing Center for Research and Education (ACCRE) at Vanderbilt University, Nashville, TN. We gratefully acknowledge the support of NVIDIA Corporation with the donation of the Titan X Pascal GPU used for this research. The imaging dataset(s) used for the analysis described were obtained from ImageVU, a research repository of medical imaging data and image-related metadata. ImageVU is supported by the VICTR CTSA award (ULTR000445 from NCATS/NIH) and Vanderbilt University Medical Center institutional funding. ImageVU pilot work was also funded by PCORI (contract CDRN-1306-04869).